\documentclass[runningheads]{llncs}
\usepackage[T1]{fontenc}
\usepackage{graphicx}
\usepackage{booktabs}
\usepackage[misc]{ifsym}

\usepackage{blindtext}
\usepackage{balance}
\usepackage{subfigure}
\usepackage{dsfont}
\usepackage{amssymb}
\usepackage{amsfonts}
\usepackage{bbm}

\newcommand{\corr}{(\Letter)}
\usepackage{mwe}

\begin{document}


\title{A Multi-class Ride-hailing Service Subsidy System Utilizing Deep Causal Networks\thanks{This paper has been accepted by ECML-PKDD 2024 (Industry Track). }}

\titlerunning{ }

\author{Zhe Yu \and
Chi Xia \and
Shaosheng Cao \corr \and 
Lin Zhou }

\authorrunning{Yu et al.}


\institute{DiDi Chuxing \\ \email{ \{yzae2623, shelsoncao\}@gmail.com, \{xiachi, realzhoulin\}@didiglobal.com} }


\maketitle              

\begin{abstract}
In the ride-hailing industry, subsidies are predominantly employed to incentivize consumers to place more orders, thereby fostering market growth. Causal inference techniques are employed to estimate the consumer elasticity with different subsidy levels. However, the presence of confounding effects poses challenges in achieving an unbiased estimate of the uplift effect. We introduce a consumer subsidizing system to capture relationships between subsidy propensity and the treatment effect, which proves effective while maintaining a lightweight online environment.



\end{abstract}

\section{Introduction}

For online ride-hailing services, a pivotal marketplace strategy involves dynamically balancing demand and supply through the judicious use of subsidies for both consumers and drivers \cite{bimpikis_spatial_2019,Castillo&Etal:2017ACM,Chen&Etal:2015ACM,shu2022modeling,wu_spatial_2020,Yan&Etal:2020NRL,yang2022grflift}. It is common practice to distribute subsidies to consumers upon receiving queries, which include the origin, destination, desired departure time, and other pertinent information. Subsequently, consumers make decisions on whether to proceed with the payment process, taking into account the subsidy offered to them. The overarching marketplace strategy aims to leverage a limited financial budget to expand the market scale.

Causal inference provides a fundamental approach to modeling consumer subsidy elasticity, which enables optimal budget allocation decisions \cite{Cohen&Etal:SSEP,yu2023consumer}. In this context, the various subsidy levels are the treatment, and the outcome is the consumer order placements. This outcome is quantified through the conversion rate, which is calculated as the ratio of the number of orders to the number of queries.
Causal inference aims to estimate the treatment effect, also known as the uplift effect, in comparison to the control group. 
Ideally, the model should be trained on data from Randomized Control Trials (RCTs)\cite{kohavi2013online,kohavi2009controlled}, where subsidies are randomly distributed to queries. However, RCTs are often impractical or too costly to conduct. 
Conversely, there is an abundance of observational data biased by the human-driven subsidy strategy, where confounding factors frequently exist in the input features, simultaneously influencing both the treatment and potential outcomes. For instance, subsidies are typically given to low-active or inactive users, which can mislead models because the conversion rate of consumers who receive subsidies is usually much lower than that of consumers who do not receive subsidies in the observational data.
Several methods have been devised to mitigate the bias introduced by confounders, where sensitivity analyses \cite{rosenbaum1983central} fall short of providing precise point estimations, and instrumental variable approaches \cite{angrist1996identification} only partially identify conditional average treatment effect. Moreover,  most existing methods cannot handle multi-categorical or continuous subsidy levels.

With the evolution of online ride-hailing platforms, consumers are presented with a range of multi-class service options for the same trip. For example, Uber offers services like Comfort, UberX, and Uber Black for identical trips, each with varying costs and subsidy levels, analogous to multi-class airline seating. Given these diverse service and subsidy options, consumers may exhibit varying consumption preferences. When a consumer places an order for multiple services, the platform searches for service suppliers, and ultimately, only one service is provided to the consumer with a certain probability. In this paper, we introduce a multi-class ride-hailing subsidy system that has been deployed in a real-world environment. A \textbf{Mul}ti-class \textbf{T}r\textbf{e}atment Generalized Propensity Score \textbf{Net}work (MulTeNet) is introduced for modeling consumer subsidy elasticity, and a Model Predictive Control (MPC) framework is applied to optimize budget distribution.

\section{The Proposed Method and Deployed System}

Given a total financial budget, the objective is to identify an optimal policy for subsidizing queries to maximize overall revenue\footnote{Due to limited space, the incentive for drivers is not disscussed. }.

\subsection{Multi-class Treatment Generalized Propensity Score Network}\label{sec:GPS}

\vspace{-7mm}
\begin{figure}[h!]
      \centering
      \includegraphics[width=0.55\textwidth]{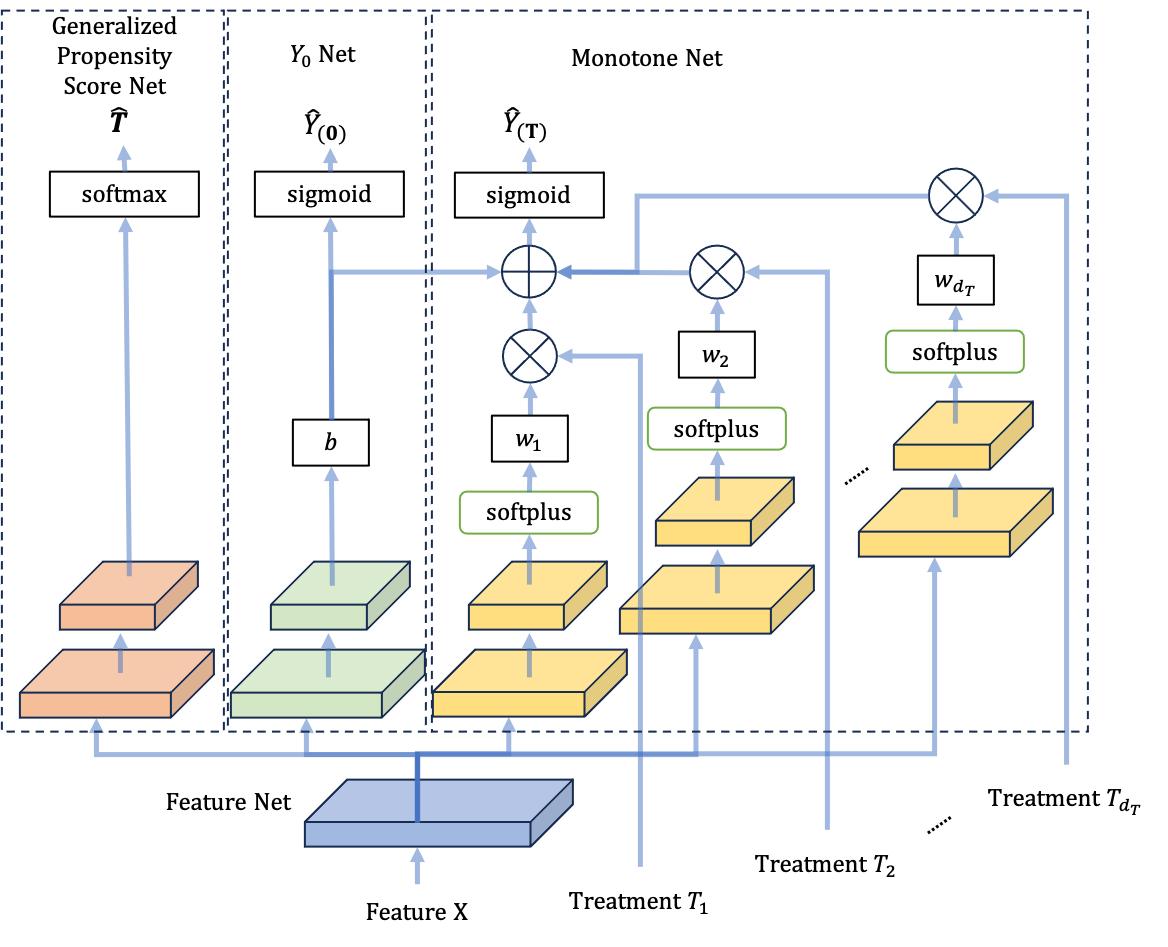}
      \vspace{-5mm}
      \caption{The illustrated structure of the proposed MulTeNet}
      \label{fig:DONUT}
\end{figure}
\vspace{-6mm}

As shown in Fig. \ref{fig:DONUT}, the model is composed of three major components. Input features excluding treatments are processed by a common feature net, and the output is then directed to three distinct networks. The Generalized Propensity Score Net, which is a multi-class neural network, estimates the probability density function of treatment. The $Y_0$ Net takes the output from the feature net and estimates the base conversion rate without any treatment. The Monotone Net estimates the uplift effect of multi-class treatments. To ensure the monotonicity of the treatment effect, the softplus activation function is utilized. Inspired by \cite{hatt2021estimating}, orthogonal regularization is introduced to rectify any biases present in the observational data.

\subsection{Optimization Formulation}

Similar to \cite{yu2023consumer}, the query-wise elasticity, as inferred by the proposed MulTeNet model, is then clustered by the origins, destinations and departure time. We formulate the budget distrubtion problem as follows:
\begin{small}
\begin{equation}
\label{eqn:MPC}
\begin{array}{ll}
\max_{x_{i,j}} &\sum_{k}\gamma_k\sum_i \hat{N}_{i}\hat{Pr}_{k,i}\sum_j\hat{P}_{i, j}x_{i,j}\\
{\mbox{s.t.}}&\sum_{k}\gamma_k\sum_i \hat{N}_{i}\sum_j\hat{P}_{i, j}x_{i,j}c_{k,i, j}\le B\\
&0\le\underline{u}\le \sum_j\hat{P}_{i, j}x_{i,j}c_{k,i, j}\le \bar{u}, \forall i, k\\
&\sum_jx_{i,j}=1, \forall  i \\
&x_{i,j}\in\{0, 1\}
\end{array}
\end{equation}
\end{small}
where $k$ is the service class index, $i$ is the cluster index, and $j$ enumerates the multi-class treatment space. 
${x_{i,j}}$ is a binary control variable indicating the $j$-th subsidy is allocated to cluster $i$. 
$\hat{P}_{i,j}$ is the estimated conversion rate for cluster $i$ given the $j$-th treatment level, 
$\hat{Pr}_{k, i}$ is the predicted order revenue for service class $k$ in cluster $i$, 
$c_{k, i, j}$ is the financial cost of the $j$-th subsidy level, 
$\gamma_k$ is the statistic probability that service class $k$ is completed by supplyers, 
$\hat{N}_{i}$ is the predicted order number, and 
$B$ is the budget limit. 
The subsidy levels u are bounded by lower and upper limits 
$\underline{u}$ and $\bar{u}$ respectively. 
To efficiently solve the large-scale linear programming, a primal-dual approximation \cite{hao2020dynamic} is employed. Additionally, to ccount for modeling errors and randomness, MPC technology \cite{qin2003survey} is utilized at regular intervals with updated information.

\subsection{System Implementation}

\vspace{-6.5mm}
  \begin{figure}[!htbp]
      \centering
      \includegraphics[width=0.65\textwidth]{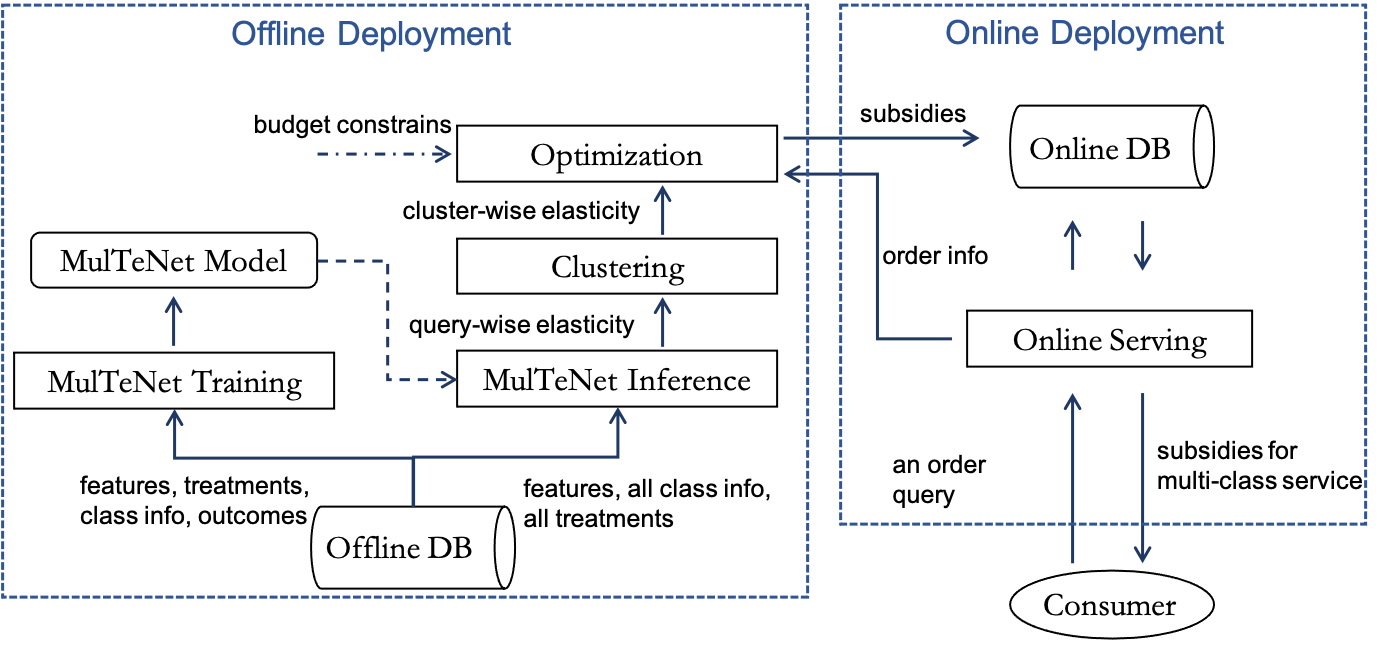}
      \vspace{-3mm}
      \caption{The framework of our system}
      \label{optimization_Framework}
   \end{figure}
 \vspace{-6mm}

As depicted in Fig. \ref{optimization_Framework}, our system encompasses both offline and online components. The MulTeNet model is trained using historical features and treatment information for different service classes. Subsequently, the model is deployed to infer query-wise elasticity at regular intervals. A clustering procedure is implemented to ensure consumer fairness. The cluster-wise elasticity and budget constraints are then fed into the optimization module, which generates a subsidy allocation dictionary for each service class. In the online phase, when a consumer inputs trip information, queries for different services are sent to the online server. The server simply looks up the dictionary and returns the corresponding results, significantly reducing the computational load in the online environment.

\section{Empirical Experiments}

\vspace{-7mm}
\begin{table}[ht]
\centering
\caption{\label{tab:demand_curve_result} Casual inference results}
\vspace{-1mm}
\resizebox{0.73\textwidth}{!}
{
\begin{tabular}{ | c | c | c | c |} 
\hline
 Model & AUC  & AUUC & QINI Coefficient\\ \hline
 Yu\&Etal2023   &   \textbf{0.8065$\pm$ 0.0001}       &   1.8086$\pm$ 0.0879   &      0.2155$\pm$ 0.0023     \\ \hline
 DragonNet   &   0.8039$\pm$ 0.0002        &   1.3825$\pm$ 1.0462   &      0.2016$\pm$ 0.0594    \\ \hline
MulTeNet   &  0.7875$\pm$ 0.0005       &  \textbf{1.9363$\pm$ 0.2135}   & \textbf{0.3249$\pm$ 0.0250}  \\  \hline
\end{tabular} 
}
\end{table}
\vspace{-5mm}

Empirical experiments were conducted to evaluate the performance of the proposed system. The dataset, comprising approximately 2 million queries with 200,000 subsidies, was extracted from a major ride-hailing platform in China between September and November 2023. For comparison, the baseline model \texttt{Yu\&Etal2023} employs a transfer learning approach \cite{yu2023consumer}. Additionally, the DragonNet model, which is a multi-task DNN predicting potential outcomes and determining subsidy allocation, was also assessed. As shown in Table \ref{tab:demand_curve_result}, our MulTeNet model achieves the highest AUUC (Area Under the Uplift Curve) and QINI coefficient, indicating its superior effectiveness in modeling uplift effects within the data. However, the AUC for MulTeNet is slightly lower, attributable to the introduction of treatment loss and regularization, which may impact response performance.

\vspace{-3mm}
\begin{table}[ht]
\centering
\caption{\label{tab:synthetic_result} Optimization results}
\vspace{-2mm}
\resizebox{0.65\textwidth}{!}
{
\begin{tabular}{ | c | c | c | c | c | c |} \hline
                & Yu\&Etal2023  & DragonNet & MulTeNet     &  No Subsidy \\ \hline
 \mbox{Revenue} & 1.0394        & 1.0283    & \textbf{1.0547}     & 1.0\\ \hline
 \mbox{Orders}  & 1.0224        & 1.0155    & \textbf{1.0298}     & 1.0 \\ \hline
 Subsidy Rate   & 4.84\%        & 4.88\%    & \textbf{4.93\%}     & - \\ \hline
 ROI            & 0.79          & 0.56      & \textbf{1.05}       & - \\ \hline
\end{tabular} }
\end{table}
\vspace{-5mm}

The experiment on various subsidy strategies was conducted over a week in December 2023, with the target subsidy rate set to 5\%. The 14-day historical data were used to forecast the subsequent 7 days, and the historical subsidy outcomes were fed back into the optimization solver daily to adjust the subsidy allocation for the remainder of the week. The ROI (Return On Investment) is defined as the increase in revenue over the total subsidy cost. As illustrated in Table \ref{tab:synthetic_result}, our system successfully controlled the subsidy rate to 4.93\%, which is closer to the target level. More importantly, The total orders and revenue of the platform increased by 5.47\% and 2.98\%, respectively, and the ROI was 1.05, which is 34\% higher than the best baseline.

\end{document}